\documentclass[10pt,conference]{IEEEtran}
\usepackage{multirow}

\usepackage{graphicx}   % Required for including images
\usepackage{subcaption} % Required for subfigures
\usepackage{float}      % Optional: for better control of figure placement

\usepackage{cite}
\usepackage{amsmath,amssymb,amsfonts}
\usepackage{tabularx}
\usepackage{array}
\usepackage{algpseudocode}
\usepackage{algorithm}
\usepackage{stfloats}
\usepackage{textcomp}
\usepackage{booktabs}
\usepackage{balance}

\def\BibTeX{{\rm B\kern-.05em{\sc i\kern-.025em b}\kern-.08em
    T\kern-.1667em\lower.7ex\hbox{E}\kern-.125emX}}

\begin{document}
\author{
    \IEEEauthorblockN{Alberto Huertas Celdrán\IEEEauthorrefmark{1}, Chao Feng\IEEEauthorrefmark{1}, Sabyasachi Banik\IEEEauthorrefmark{1}, G\'er\^ome Bovet\IEEEauthorrefmark{2}, \\Gregorio Mart\'inez P\'erez\IEEEauthorrefmark{3}, Burkhard Stiller\IEEEauthorrefmark{1}}
    
    \IEEEauthorblockA{\IEEEauthorrefmark{1}Communication Systems Group CSG, Department of Informatics, University of Zurich UZH, CH--8050 Zürich, Switzerland \\{[cfeng, huertas, stiller]}@ifi.uzh.ch, sabyasachi.banik@uzh.ch}
    \IEEEauthorblockA{\IEEEauthorrefmark{2}Department of Information and Communications Engineering, University of Murcia, 30100--Murcia, Spain {gregorio}@um.es}
    \IEEEauthorblockA{\IEEEauthorrefmark{2}Cyber-Defence Campus, armasuisse Science \& Technology, CH--3602 Thun, Switzerland gerome.bovet@armasuisse.ch}
}

% change symbols of affiliations to numbers
\DeclareRobustCommand*{\IEEEauthorrefmark}[1]{%
  \raisebox{0pt}[0pt][0pt]{\textsuperscript{\footnotesize #1}}%
}

% \title{\solution{}: MTD-Based Mitigation of Poisoning Attacks on Decentralized Federated Learning}
\title{De-VertiFL: A Solution for Decentralized \\ Vertical Federated Learning}

\maketitle

\begin{abstract}
Federated Learning (FL), introduced in 2016, was designed to enhance data privacy in collaborative model training environments. Among the FL paradigm, horizontal FL, where clients share the same set of features but different data samples, has been extensively studied in both centralized and decentralized settings. In contrast, Vertical Federated Learning (VFL), which is crucial in real-world decentralized scenarios where clients possess different, yet sensitive, data about the same entity, remains underexplored. Thus, this work introduces De-VertiFL, a novel solution for training models in a decentralized VFL setting. De-VertiFL contributes by introducing a new network architecture distribution, an innovative knowledge exchange scheme, and a distributed federated training process. Specifically, De-VertiFL enables the sharing of hidden layer outputs among federation clients, allowing participants to benefit from intermediate computations, thereby improving learning efficiency. De-VertiFL has been evaluated using a variety of well-known datasets, including both image and tabular data, across binary and multiclass classification tasks. The results demonstrate that De-VertiFL generally surpasses state-of-the-art methods in F1-score performance, while maintaining a decentralized and privacy-preserving framework.\end{abstract}


\section{Introduction}
\label{intro}

Artificial Intelligence (AI), and particularly Machine Learning (ML), is entering a new era due to the exponential increase in data generated by the vast number of Internet of Things (IoT) devices \cite{explodingtopics2024}. This surge in data offers a rich environment for the rapid evolution and refinement of ML models. However, traditional ML approaches, which require aggregating data from multiple sources into a central repository, carry significant risks, such as data breaches and privacy violations. In 2016, Google proposed Federated Learning (FL) \cite{FL} as a solution to enable collaborative model training without data centralization. Since that, FL has found widespread application in critical areas such as healthcare, finance, and autonomous driving.

Among the various FL scenarios, Horizontal Federated Learning (HFL) \cite{HFL} has been the most extensively studied in the literature. In HFL, all clients in the federation hold datasets with the same feature set but different samples. Additionally, while some research has explored decentralized implementations of HFL, the majority focuses on centralized approaches \cite{DFL}. In this context, Decentralized Federated Learning (DFL) aims to address issues, such as single points of failure and bottlenecks, found in centralized FL. However, FL extends beyond HFL, and Vertical Federated Learning (VFL) \cite{VFL} is an interesting and promising avenue for innovation. VFL is particularly beneficial when multiple clients possess complementary features about the same entities, enabling more complex and insightful learning models.

Despite the potential of VFL, previous research has predominantly focused on centralized approaches for binary classification, often limited to cases involving just a few participants \cite{VL2}. This indicates a clear need for further investigation into fully decentralized VFL systems that can accommodate multiple participants. This gap, alongside challenges such as efficient model synchronization, knowledge exchange, and maintaining model performance in a decentralized setting, are essential for developing more adaptable and robust VFL solutions suitable for real-world scenarios. 

Therefore, this work present De-VertiFL (publicly available in \cite{banik_decentralizedvfl}), a solution for training FL models in a vertical a decentralized fashion. The proposed solution proposes novel architectural decisions, an innovative knowledge exchange scheme, and a distributed federated training process. More in particular, it introduces a novel model training through the sharing of hidden layer outputs and gradient exchange in a peer-to-peer manner. Extensive experiments were conducted to evaluate De-VertiFL performance, using different number of participants in the VFL setup. Furthermore, the evaluation was performed on several benchmark datasets, including MNIST, FMNIST, Titanic, and Bank Marketing. Key insights were gained regarding how the number of participants and the complexity of the neural network architecture affect learning outcomes and model performance. Finally, De-VertiFL was compared to state-of-the-art methods, demonstrating versatility across both multi-class and binary classification tasks.

The remainder of this paper is structured as follows. Section~\ref{sec:relatedwork} reviews related literature. Section~\ref{sec:solution} delves into the architectural design and implementation of the solution, and Section~\ref{sec:eval} evaluates experimental results. Finally, Section~\ref{sec:conclusions} outlines conclusions and future research. 
\section{Related Work}
\label{sec:relatedwork}

This section reviews studies on VFL in centralized and decentralized approaches. It first examines binary classification studies, which inform the exploration of VFL for multi-class classification.

\cite{VL2} proposed a binary classification approach using Asymmetric Private Set Intersection (APSI) and Asymmetric Vertical Logistic Regression (AVLR) in a centralized setting. Despite its focus on two-party scenarios and customized strict ID intersection, the study demonstrated comparable performance to symmetric versions, such as AUC and loss metrics, validating APSI and AVLR's feasibility. \cite{VL3} introduced Private Decision Logistic Regression (PDLR) for centralized VFL, highlighting data leakage risks and concerns about its applicability in decentralized frameworks. Similarly, \cite{VL5} explored Gradient Boosting Decision Trees (GBDT) for binary classification, emphasizing its efficiency and scalability while noting the limitations of tree-based methods for complex tasks, advocating for DNN-based approaches.

Federated Deep Support Kernel Learning (FDSKL) \cite{VL8} employed kernel methods for binary classification, outperforming state-of-the-art methods in speed and generalization. A Flower-based VFL solution applied to the Titanic dataset demonstrated efficient three-client binary classification using neural network embeddings and centralized gradient aggregation.

For multi-class classification, \cite{VL10} proposed Multi-View Vertical Federated Learning (MMVFL), combining unsupervised label sharing and multi-view learning in a centralized framework. \cite{VP1} presented PyVertical, a centralized framework enabling learning without data sharing through Private Set Intersection (PSI). Both approaches raised concerns about centralization, suggesting further exploration of decentralized techniques.

The only decentralized VFL approach, \cite{sanchez2024analyzing}, introduced VertiChain and VertiComb, focusing on privacy-preserving neural network training for vertical scenarios. Experiments assessed their performance under non-IID data and adversarial attacks, emphasizing robustness.

In conclusion, research predominantly focuses on centralized binary classification with limited decentralization. Expanding decentralized VFL frameworks to handle multi-class classification and complex feature distributions remains a critical research direction.

\section{De-VertiFL}
\label{sec:solution}

This section outlines the design and implementation of De-VertiFL, a decentralized framework for vertical federated learning (VFL).

\begin{figure}[htpb!]
    \centering
    \includegraphics[width=1\columnwidth]{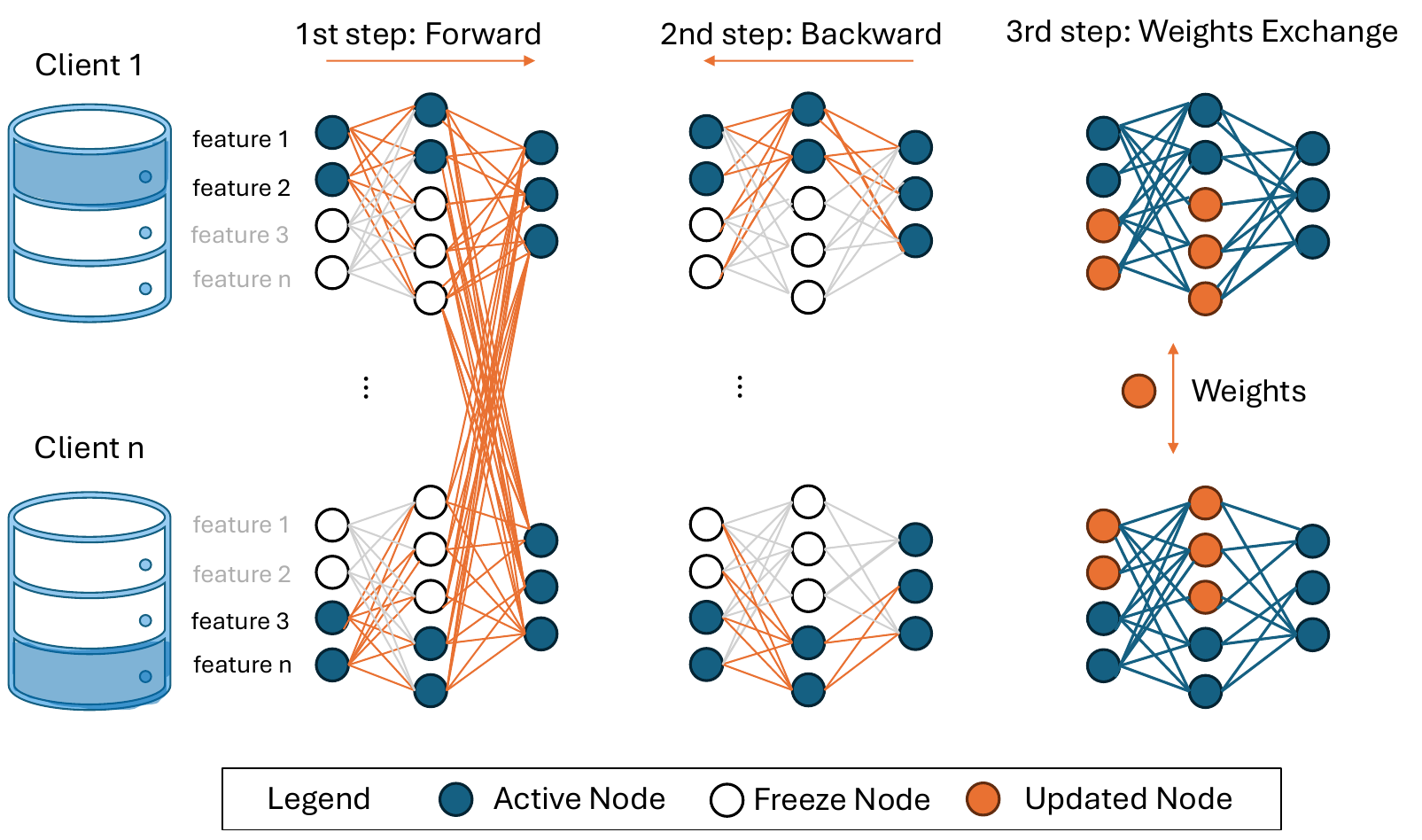}
    \caption{De-VertiFL Architectural Design}
    \label{fig:architecture}
\end{figure}

At the start of training, clients collaboratively decide on a neural network architecture based on the task and dataset, as if all features were centralized. Each client is assigned a specific feature set and corresponding neural network segment (input and hidden layers), determined by vertical data partitioning (\figurename~\ref{fig:architecture}, left). Clients process their local data using zero-padding for missing features, ensuring uniform input sizes. During the forward pass (step 1), padded inputs activate relevant network components, with participants sharing hidden layer outputs (in orange, \figurename~\ref{fig:architecture}). This intermediate exchange improves accuracy and robustness despite fragmented features.

In the backward pass (step 2), clients update weights only for their assigned model segments. These weights are then shared peer-to-peer (step 3), bypassing the need for a central server. Participants aggregate received weights using the FedAvg function, synchronizing global model updates.

This process, repeated over multiple rounds, enables the global model to learn progressively from distributed datasets. Each federated round involves local training, weight exchange, and aggregation in a decentralized manner.

% Implementation
The De-VertiFL algorithm has been implemented using PyTorch. Data preprocessing and handling have been conducted using Pandas and NumPy for dataset manipulation. For image datasets, the torchvision library is employed for data loading and transformations. For tabular datasets, scikit-learn is used for tasks such as scaling, one-hot encoding, and train-test splitting. In addition, the models are based on multi-layer perceptrons (MLPs) with fully connected layers, ReLU activation functions, and dropout layers for regularization. Training is optimized using the Adam optimizer, and either CrossEntropyLoss or BCEWithLogitsLoss is applied depending on the classification task. Model performance is evaluated using accuracy and F1-score metrics, with results visualized using Matplotlib to track performance over federated rounds. Confidence intervals and error bars are generated using the SciPy sublibrary `scipy.stats`, which computes and visualizes performance variability.

\section{Experiments}
\label{sec:eval}

This section presents a pool of experiments to measure the performance of De-VertiFL on heterogeneous datasets (containing images and tabular data) and compare the results with non-federated scenarios and the state of the art. The datasets used for evaluation are: MNIST, Fashion-MNIST, Titanic, and Bank Marketing.

\subsection{MNIST Dataset}

The MNIST dataset comprises 70,000 grayscale images of handwritten digits (28x28 pixels each), with 784 features (pixels) ranging from 0 to 255. De-VertiFL distributes features among participants in a round-robin manner. For instance, in a 7-participant setup, each participant receives 112 features (4 rows), while in a 2-participant setup, each receives 392 features (14 rows).

Experiments involve 2 to 10 clients using a multi-layer perceptron for classification. The input layer size depends on the number of features per participant, while the network includes 3 hidden layers (10 neurons each) and an output layer with 10 neurons for the 10 MNIST classes. Training spans 5 epochs per local session across 5 federated rounds, averaging results over 5 seeds for reliability.

De-VertiFL’s performance is benchmarked against a non-federated baseline, where participants train independently without knowledge exchange. This baseline reflects individual model performance. \tablename~\ref{tab:performance} compares the F1-scores of De-VertiFL and non-federated setups across varying client configurations, with best-performing results highlighted in \textbf{bold}.

As shown in \tablename~\ref{tab:performance}, even with just two participants, De-VertiFL significantly outperforms the non-federated scenario. However, as the number of participants increases, the F1-score for both methods decreases due to fewer features per client, reducing available knowledge. This impact is more pronounced in non-federated approaches, lacking collaboration. For instance, with nine clients, De-VertiFL achieves an F1-score of 68\%, compared to only 13\% in the non-federated scenario, highlighting the benefits of knowledge exchange in De-VertiFL.

\subsection{FMNIST Dataset}

The Fashion-MNIST dataset contains 70,000 grayscale annotated fashion images, each measuring 28x28 pixels. The dataset is divided into 60,000 training images and 10,000 testing images. Each image depicts a fashion item, such as clothing, footwear, or accessories, from 10 different classes: T-shirt/top, Trouser, Pullover, Dress, Coat, Sandal, Shirt, Sneaker, Bag, and Ankle boot.

As in the previous experiment, the task is to classify these 10 classes, and the federation setup ranges from 2 to 10 clients. The features are distributed across the clients in a similar manner as in the previous experiment, and the same neural network architecture and training configuration (epochs and rounds) are used. \tablename ~\ref{tab:performance} shows the F1-score of De-VertiFL compared to the non-federated scenario for different federation configurations.

As shown in \tablename ~\ref{tab:performance}, De-VertiFL demonstrates strong and stable performance, achieving $>$75\% of F1-score with up to seven participants. It is worth noting that as the number of participants increases to eight or more, the F1-score decreases to around 65\%. In comparison, De-VertiFL significantly outperforms the non-federated setup, where performance steadily declines as the number of participants increases, dropping to 16\% F1-score with 10 participants.

\subsection{Titanic Dataset}

The Titanic dataset contains 891 records with features like passenger name, age, gender, ticket class, fare, and survival status (label). The task is to predict passenger survival based on these features.

During preprocessing, irrelevant features (\textit{PassengerId}, \textit{Name}, \textit{Ticket}) were removed. Common techniques included age binning (grouping \textit{Age} into categories like Child, Adult, Elderly, Unknown), cabin extraction (using the first letter of \textit{Cabin} and treating missing values as \textit{Unknown}), and title extraction (creating a new feature by simplifying titles from \textit{Name}). The final features were: \textit{Pclass}, \textit{Sex}, \textit{Age}, \textit{SibSp}, \textit{Parch}, \textit{Fare}, \textit{Cabin}, \textit{Embarked}, and \textit{Title}. For federation configurations (2–9 participants), features were randomly distributed among clients. The same neural network architecture and training setup from prior experiments were used, with the input layer adjusted per client’s features and a two-node output layer. Training involved 1000 federated rounds with 1 epoch per round. \tablename~\ref{tab:performance} compares De-VertiFL’s F1-score to the non-federated scenario across configurations.

As shown in \tablename~\ref{tab:performance}, De-VertiFL consistently outperforms the non-federated version across all configurations, with F1-scores over 30\% higher. For example, with nine participants, De-VertiFL achieves 76\% compared to 41\% for the non-federated approach. This highlights the effectiveness of its knowledge-sharing mechanism. Additionally, De-VertiFL maintains stable F1-scores as participant numbers grow, due to the binary nature of the Titanic survival task, which is simpler than the multi-class classification problems in MNIST and FMNIST. The reduced feature space and simpler decision boundaries mitigate performance degradation.

\subsection{Bank Marketing Dataset}

The Bank Marketing dataset includes information on Portuguese bank marketing campaigns, aiming to predict if clients will subscribe to a term deposit. Features include \textit{Age}, \textit{Job}, \textit{Marital status}, \textit{Education}, \textit{Default}, \textit{Balance}, \textit{Housing}, \textit{Loan}, \textit{Contact}, \textit{Day}, \textit{Month}, \textit{Campaign}, \textit{Pdays}, \textit{Previous}, and \textit{Poutcome}. This is a binary classification task.

Preprocessing involved removing irrelevant columns like \textit{Duration}, encoding binary features (\textit{Default}, \textit{Housing}, \textit{Loan}, \textit{Deposit}), and applying one-hot encoding to other categorical features (\textit{Job}, \textit{Marital status}, \textit{Education}, etc.). Features were normalized using StandardScaler. Features were distributed among participants per federation setup. The same neural network architecture as previous experiments was used, with input layers adjusted to assigned features and a two-node output layer for binary classification. Training consisted of 20 federated rounds with 10 epochs each. \tablename~\ref{tab:performance} compares De-VertiFL's F1-scores to the non-federated scenario across configurations.

As shown in \tablename~\ref{tab:performance}, De-VertiFL outperforms non-federated scenarios in all configurations, maintaining a stable F1-score of 67\% as the number of clients increases. This stability reflects the binary classification task's lower variability compared to multi-class problems. The non-federated approach also shows a flat performance curve, with F1-scores decreasing slightly from 65\% to 56\% as participants increase, highlighting its robustness but lack of collaborative benefits. These results emphasize De-VertiFL's advantage in improving prediction accuracy through client collaboration.

\begin{table*}[ht]
\caption{F1-Score of De-VertiFL and Comparison with a Non-federated Approach and Literature.}
\centering
\begin{tabular}{cc|ccccccccc}
\toprule
\multirow{2}{*}{\textbf{Dataset}} & \multirow{2}{*}{\textbf{Approach}} & \multicolumn{9}{c}{\textbf{F1-Score According to the Number of Participants}} \\ \cline{3-11}
 &  & \textbf{2} & \textbf{3} & \textbf{4} & \textbf{5} & \textbf{6} & \textbf{7} & \textbf{8} & \textbf{9} & \textbf{10} \\ \hline
 \multirow{3}{*}{MNIST} & De-VertiFL & \textbf{0.960}  & \textbf{0.955}  & \textbf{0.942}  &  \textbf{0.910} & \textbf{0.870}  & \textbf{0.822}  & \textbf{0.764}  & \textbf{0.684}  &  \textbf{0.584}  \\ \cline{2-11}
                           & Non-federated & 0.822  &  0.619 &  0.480 & 0.345  & 0.261  &  0.159 & 0.165  & 0.130  &  0.100
                           \\ \cline{2-11}
                           & VertiComb & 0.932  & 0.902  & 0.866   & 0.838  & 0.746  & 0.670  &  0.664 & 0.544  & 0.450  \\ \hline \hline
\multirow{3}{*}{FMNIST} & De-VertiFL & \textbf{0.824}  & \textbf{0.812}   & \textbf{0.810}  & \textbf{0.788}  & \textbf{0.784}  & \textbf{0.764}  &  0.688 &  0.652 & \textbf{0.648}   \\ \cline{2-11}
                           & Non-federated & 0.753  &  0.658 & 0.617  &  0.548 & 0.424  & 0.284  & 0.257  &  0.211 &  0.159  \\ \cline{2-11}
                           & VertiComb &  0.816 & 0.798  & 0.780 &  0.748 & 0.740 & 0.734  & \textbf{0.714} &  \textbf{0.712} & 0.628 \\ \hline \hline

\multirow{3}{*}{Titanic} & De-VertiFL & \textbf{0.770}  & \textbf{0.762}  &  0.779 & \textbf{0.781}  &  \textbf{0.713} &  \textbf{0.714} & 0.719  & \textbf{0.760}  &  -  \\ \cline{2-11}
                           & Non-federated & 0.507  & 0.497  & 0.460  & 0.441  & 0.426  & 0.445  &  0.403 &  0.411 &   - \\
                           \cline{2-11}
                           & VertiComb & 0.516  & 0.530  & \textbf{0.792}  & 0.708  & 0.471  & 0.712  & \textbf{0.728}  &  0.707 & - \\ 
                           \hline \hline
\multirow{2}{*}{Bank Marketing} & De-VertiFL & \textbf{0.683}  &  0.667 & \textbf{0.680}  & 0.667  & \textbf{0.681}  & \textbf{0.681}  &  \textbf{0.679} &  0.676 & \textbf{0.675}  \\ \cline{2-11}
                           & Non-federated & 0.647  & 0.630  & 0.614  &  0.577 & 0.554  & 0.581  & 0.574  &  0.583 &  0.564  \\ \cline{2-11}
                           & VertiComb & 0.678  & \textbf{0.683} &  0.669 &  \textbf{0.678} & 0.651 &  0.667 & 0.658 &  \textbf{0.677} & 0.666 \\ \bottomrule
\end{tabular}
\label{tab:performance}
\end{table*}

\subsection{Comparison with Literature}

The most closely related approach to De-VertiFL is VertiComb, introduced in \cite{sanchez2024analyzing}, which also adopts a vertical and decentralized framework for federated learning. Therefore, VertiComb's code was re-implemented and executed under the same participant setup described earlier. \tablename ~\ref{tab:performance} illustrates the F1-Score comparison between De-VertiFL and VertiComb. As can be seen in De-VertiFL outperforms VertiComb in almost all datasets and participant configurations. These improvement relies on the novelty provided by De-VertiFL in terms of knowledge exchange during the forward process, instead of the backward.

% Other comparisons

In addition, other frameworks such as PyVertical~\cite{VP1}, Flower~\cite{flower}, and SplitNN~\cite{splitNN} have also evaluated their performance using the MNIST, Titanic, or Bank Marketing datasets, albeit with fixed configurations in terms of participants.

For the MNIST dataset, PyVertical trained a neural network in centralized, federated, and vertically partitioned settings, with data split across 2 clients and 5 federated rounds following 5 initial iterations. Flower evaluated the Titanic dataset, vertically partitioned among 3 participants, over 1,000 federated rounds with 1 epoch per round. SplitNN used the Bank Marketing dataset, distributing data between 2 participants and training over 20 federated rounds with 20 local epochs each. To ensure a fair comparison, De-VertiFL was tested under the same configurations as PyVertical, Flower, and SplitNN. \tablename~\ref{tab:comparison} shows the performance comparison across these frameworks.

\begin{table}[ht]
\centering
\caption{Accuracy of De-VertiFL and Related Work}

\begin{tabular} {llll}
\hline
\textbf{Approach} & \textbf{Performance (\%)} & \textbf{Dataset}  & \textbf{Participants} \\ \hline \hline

De-VertiFL & $\approx$96\% Accuracy & MNIST & 2  \\ \hline
PyVertical \cite{VP1} & $\approx$93\% Accuracy & MNIST & 2  \\ \hline \hline

De-VertiFL & $\approx$80\%  Accuracy & Titanic & 3  \\ \hline
Flower \cite{flower} & $\approx$65\% Accuracy & Titanic & 3  \\ \hline \hline

De-VertiFL & $\approx$70\% F1-Score & Bank Marketing & 2  \\ \hline
SplitNN \cite{splitNN} & $\approx$47\% F1-Score & Bank Marketing & 2  \\ \hline

\end{tabular}

\label{tab:comparison}
\end{table}

As shown in \tablename~\ref{tab:comparison}, De-VertiFL consistently outperforms the three related works across different datasets. In conclusion, De-VertiFL outperforms existing methods in both image and binary classification tasks, including MNIST, FMNIST, Titanic, and Bank Marketing, demonstrating its superiority.

\section{Conclusion and Future Work}
\label{sec:conclusions}

This work has designed and implemented De-VertiFL, a solution to train vertical FL models in an decentralized, privacy-preserving, and collaborative fashion among multiple participants, each holding distinct subsets of data features. De-VertiFL eliminates the need for a central server, thereby reducing potential single points of failure and enhancing data privacy. De-VertiFL was benchmarked against several state-of-the-art datasets, including MNIST, FMNIST, Titanic, and Bank Marketing. A pool of experiments has demonstrated the effectiveness of De-VertiFL across all datasets. In general, De-VertiFL outperformed or matched the performance of state-of-the-art centralized frameworks. More in detail, performance degradation was observed with increasing participants, primarily due to the more restricted view each participant had of the overall dataset. However, the introduction of hidden layer output and loss sharing helped mitigate some of these challenges.

As future work it is planned to study how to reduce the performance degradation as the number of participants increased. Novel strategies to enhance model performance even when the number of participants is large, potentially through more sophisticated aggregation methods or advanced network structures will be considered.

\section*{Acknowledgments}
This work has been partially supported by \textit{(a)} the Swiss Federal Office for Defense Procurement (armasuisse) with the CyberMind project (CYD-C-2020003), \textit{(b)} the University of Zürich UZH, and \textit{c} the strategic project CDL-TALENTUM from the Spanish National Institute of Cybersecurity (INCIBE) by the Recovery, Transformation, and Resilience Plan, Next Generation EU.

\bibliographystyle{IEEEtran}  
\balance
\bibliography{references}

\begin{thebibliography}{10}
\providecommand{\url}[1]{#1}
\csname url@samestyle\endcsname
\providecommand{\newblock}{\relax}
\providecommand{\bibinfo}[2]{#2}
\providecommand{\BIBentrySTDinterwordspacing}{\spaceskip=0pt\relax}
\providecommand{\BIBentryALTinterwordstretchfactor}{4}
\providecommand{\BIBentryALTinterwordspacing}{\spaceskip=\fontdimen2\font plus
\BIBentryALTinterwordstretchfactor\fontdimen3\font minus \fontdimen4\font\relax}
\providecommand{\BIBforeignlanguage}[2]{{%
\expandafter\ifx\csname l@#1\endcsname\relax
\typeout{** WARNING: IEEEtran.bst: No hyphenation pattern has been}%
\typeout{** loaded for the language `#1'. Using the pattern for}%
\typeout{** the default language instead.}%
\else
\language=\csname l@#1\endcsname
\fi
#2}}
\providecommand{\BIBdecl}{\relax}
\BIBdecl

\bibitem{explodingtopics2024}
\BIBentryALTinterwordspacing
``Amount of data created daily,'' 2024, accessed: 27 September 2024. [Online]. Available: \url{https://explodingtopics.com/blog/data-created-daily}
\BIBentrySTDinterwordspacing

\bibitem{FL}
L.~Li, Y.~Fan, M.~Tse, and K.-Y. Lin, ``A review of applications in federated learning,'' \emph{Computers \& Industrial Engineering}, vol. 149, p. 106854, 2020.

\bibitem{HFL}
Q.~Yang, Y.~Liu, T.~Chen, and Y.~Tong, ``Federated machine learning: Concept and applications,'' \emph{ACM Transactions on Intelligent Systems and Technology}, vol.~10, no.~2, 2019.

\bibitem{DFL}
E.~T. Martínez~Beltrán, M.~Q. Pérez, P.~M.~S. Sánchez, S.~L. Bernal, G.~Bovet, M.~G. Pérez, G.~M. Pérez, and A.~H. Celdrán, ``Decentralized federated learning: Fundamentals, state of the art, frameworks, trends, and challenges,'' \emph{IEEE Communications Surveys \& Tutorials}, vol.~25, no.~4, pp. 2983--3013, 2023.

\bibitem{VFL}
Y.~Liu, Y.~Kang, T.~Zou, Y.~Pu, Y.~He, X.~Ye, Y.~Ouyang, Y.-Q. Zhang, and Q.~Yang, ``Vertical federated learning: Concepts, advances, and challenges,'' \emph{IEEE Transactions on Knowledge and Data Engineering}, vol.~36, no.~7, pp. 3615--3634, 2024.

\bibitem{VL2}
Y.~Liu, X.~Zhang, and L.~Wang, ``Asymmetrical vertical federated learning,'' \emph{ArXiv}, 2020.

\bibitem{banik_decentralizedvfl}
S.~Banik, ``Decentralizedvfl: Prototype design to train vertical federated learning models in a decentralized fashion,'' \url{https://github.com/SabyasachiBanik/DecentralizedVFL}, 2024.

\bibitem{VL3}
S.~Yang, B.~Ren, X.~Zhou, and L.~Liu, ``Parallel distributed logistic regression for vertical federated learning without third-party coordinator,'' \emph{ArXiv}, 2019.

\bibitem{VL5}
F.~Fu, Y.~Shao, L.~Yu, J.~Jiang, H.~Xue, Y.~Tao, and B.~Cui, ``Vf2boost: Very fast vertical federated gradient boosting for cross-enterprise learning,'' in \emph{Proceedings of the 2021 International Conference on Management of Data}.\hskip 1em plus 0.5em minus 0.4em\relax Association for Computing Machinery, 2021, p. 563–576.

\bibitem{VL8}
B.~Gu, Z.~Dang, X.~Li, and H.~Huang, ``Federated doubly stochastic kernel learning for vertically partitioned data,'' \emph{ArXiv}, 2020.

\bibitem{VL10}
S.~Feng, H.~Yu, and Y.~Zhu, ``Mmvfl: A simple vertical federated learning framework for multi-class multi-participant scenarios,'' \emph{Sensors}, vol.~24, no.~2, p. 619, 2024.

\bibitem{VP1}
D.~Romanini, A.~J. Hall, P.~Papadopoulos, T.~Titcombe, A.~Ismail, T.~Cebere, R.~Sandmann, R.~Roehm, and M.~A. Hoeh, ``Pyvertical: A vertical federated learning framework for multi-headed splitnn,'' \emph{ArXiv}, 2021.

\bibitem{sanchez2024analyzing}
P.~M. S{\'a}nchez~S{\'a}nchez, A.~Huertas~Celdr{\'a}n, E.~T. Mart{\'\i}nez~P{\'e}rez, D.~Demeter, G.~Bovet, G.~Mart{\'\i}nez~P{\'e}rez, and B.~Stiller, ``Analyzing the robustness of decentralized horizontal and vertical federated learning architectures in a non-iid scenario,'' \emph{Applied Intelligence}, pp. 1--17, 2024.

\bibitem{flower}
D.~J. Beutel, T.~Topal, A.~Mathur, X.~Qiu, J.~Fernandez-Marques, Y.~Gao, L.~Sani, K.~H. Li, T.~Parcollet, P.~P.~B. de~Gusmão, and N.~D. Lane, ``Flower: A friendly federated learning research framework,'' \emph{ArXiv}, 2022.

\bibitem{splitNN}
\BIBentryALTinterwordspacing
I.~Ceballos, V.~Sharma, E.~Mugica, A.~Singh, A.~Roman, P.~Vepakomma, and R.~Raskar, ``Splitnn-driven vertical partitioning,'' \emph{ArXiv}, 2020. [Online]. Available: \url{https://arxiv.org/abs/2008.04137}
\BIBentrySTDinterwordspacing

\end{thebibliography}

\end{document}